\documentclass[10pt,twocolumn,letterpaper]{article}

\usepackage{iccv}
\usepackage{times}
\usepackage{epsfig}
\usepackage{subcaption}
\usepackage{graphicx}
\usepackage{amsmath}
\usepackage{amssymb}
\usepackage{booktabs}
\usepackage{multirow}
\makeatletter
\@namedef{ver@everyshi.sty}{}
\makeatother
\usepackage{tikz}  
\usepackage{pgfplots}
\usetikzlibrary{fit}
\pgfplotsset{compat=newest}%

\usepackage[pagebackref=true,breaklinks=true,letterpaper=true,colorlinks,bookmarks=false]{hyperref}

\iccvfinalcopy 

\def\iccvPaperID{***} 

\ificcvfinal\fi

\begin{document}

\title{LEAPS: End-to-End One-Step Person Search With Learnable Proposals}


\author{Zhiqiang Dong$^{1}$, Jiale Cao$^{1,4}$, Rao Muhammad Anwer$^2$, Jin Xie$^3$,   Fahad Khan$^2$, and Yanwei Pang$^{1,4}$\\
$^1$Tianjin University~$^2$Mohamed bin Zayed University of Artificial Intelligence\\$^3$Chongqing University~$^4$Shanghai Artificial Intelligence Laboratory\\
{\tt\small \{dzq,connor,pyw\}@tju.edu.cn }\\
{\tt\small \{rao.anwer,fahad.khan\}@mbzuai.ac.ae}~ {\tt\small xiejin@cqu.edu.cn}
}

\maketitle
\ificcvfinal\thispagestyle{empty}\fi

\begin{abstract}
We propose an end-to-end one-step person search approach with learnable proposals, named LEAPS. Given a set of sparse and learnable proposals, LEAPS employs a dynamic person search head to directly perform person detection and corresponding re-id feature generation without non-maximum suppression post-processing. The dynamic person search head comprises a  detection head and a novel flexible re-id head. Our flexible re-id head first employs a dynamic region-of-interest (RoI) operation to  extract discriminative RoI features of the proposals. Then, it generates re-id features using a plain and a hierarchical interaction re-id module. To better guide discriminative re-id feature learning, we introduce a diverse re-id sample matching strategy, instead of bipartite matching in detection head. Comprehensive experiments reveal the benefit of the proposed LEAPS, achieving a favorable  performance on  two public person search benchmarks: CUHK-SYSU and PRW. When using the same ResNet50 backbone, our LEAPS obtains a mAP score of 55.0\%, outperforming the best reported results in literature by 1.7\%, while  achieving around a two-fold speedup on the challenging PRW dataset. Our source code and models will be released.
\end{abstract}

\section{Introduction}
\label{sec:intro}

Person search \cite{Xiao_OIM_CVPR_2017,Zheng_PRW_CVPR_2017,Chen_NAE_CVPR_2020} is a challenging computer vision problem, where the task is to locate and identify a given  target person from a gallery of real-world scene images, which is widely applied in many applications, such as intelligent surveillance. As a joint task of person detection and re-identification (re-id), person search 
not only requires dealing with the challenges existing in these two individual sub-tasks, but also needs to jointly optimize the diverse objectives of both sub-tasks together. 

\begin{figure}[t!]
\includegraphics[width=0.98\linewidth]{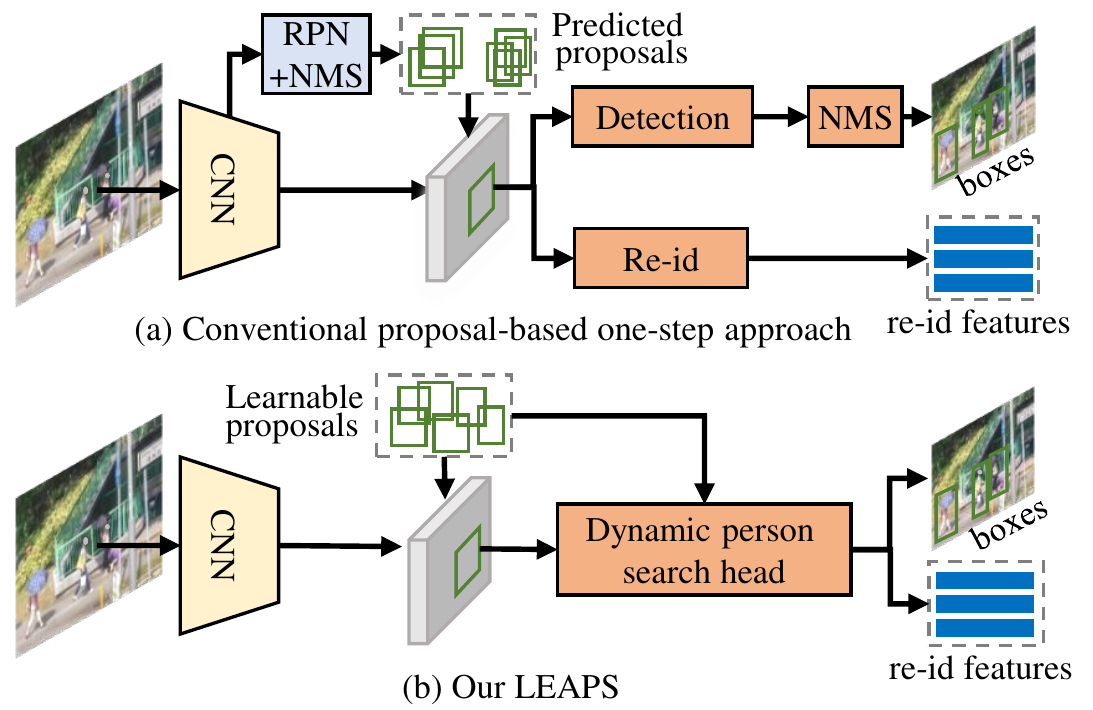} \vspace{-0.4cm}
\caption{Comparison of existing conventional proposal-based one-step approaches (a)  and our LEAPS (b). (a) The proposal-based one-step approaches \cite{Chen_NAE_CVPR_2020,Li_SeqNet_AAAI_2021,Yu_COAT_CVPR_2022} first predict person proposals from dense default anchors by RPN, and then perform detection and re-id followed by a NMS post-processing. (b) Given a set of sparse and learnable proposals, our LEAPS employs a dynamic person search head to directly perform person detection and re-id feature prediction without hand-designed anchors and NMS operation.} \vspace{-0.4cm} 
\label{fig:intro}
\end{figure}

Existing person search methods can be mainly divided into two classes: two-step and one-step approaches. Two-step approaches \cite{Chen_MGTS_ECCV_2018,Lan_CLSA_ECCV_2018,Zheng_PRW_CVPR_2017} usually adopt two independent networks for detection and re-id, respectively. In first step, they employ a detection network to detect persons from the images. Afterwards, another network is used to perform person re-id based on the  cropped persons.  Different from two-step approaches, one-step approaches \cite{Xiao_OIM_CVPR_2017,Yan_AlignPS_CVPR_2021,Cao_PSTR_CVPR_2022} aim to perform detection and re-id in a single unified network. Proposal-based one-step approaches \cite{Chen_NAE_CVPR_2020,Li_SeqNet_AAAI_2021,Yu_COAT_CVPR_2022} are one of representative methods obtaining state-of-the-art performance, which are usually built on modern object detection framework, such as Faster R-CNN \cite{Ren_FasterRCNN_NIPS_2015}. As shown in Fig. \ref{fig:intro}(a), they first predict person proposals from dense anchors with a RPN sub-network, and then perform detection and re-id, respectively. In addition, they typically employ hand-designed components, such as non-maximum suppression operation and dense default anchor settings. 


In contrast to these conventional proposal-based one-step approaches, we aim to design an end-to-end person search framework with learnable proposals, which does not require hand-designed anchors and NMS operation. In person search task, the detection sub-task aims to detect all persons in an image, while the re-id sub-task strives to distinguish different persons. Therefore, compared to detection, re-id requires more finer and discriminative person features.  We mainly  consider  two useful properties for discriminative re-id feature extraction. \textbf{(i) Rich instance information mining:} Most proposal-based one-step person search approaches typically employ RoI pooling operation followed by several fixed fully-connected layers to extract re-id features. Such a strategy does not exploit rich inter-instance and intra-instance information for re-id sub-task. To this end, a flexible re-id head is desired to effectively encode rich person instance information. \textbf{(ii) Diverse re-id samples for training:} Most end-to-end object detectors \cite{Carion_DETR_ECCV_2020,Sun_Sparse_CVPR_2021} adopt one-to-one bipartite matching strategy between predictions and ground-truths. Here, each person corresponds to one positive sample. Distinct from the person detection sub-task, we argue that the re-id sub-task requires diverse instance-level samples to learn discriminative re-id features for each person. 


\begin{figure}[t!]
\centering
\resizebox{8.0cm}{!}{
\begin{minipage}{.475\linewidth}
	\resizebox{4.2cm}{!}{%
		\begin{tikzpicture} 
		\begin{axis}[
		axis lines = left,
		ymin=20, ymax=56, 
		xmin=45, xmax=105,
		xlabel=Inference time (ms),
		ylabel= mAP,
		]		
		\coordinate (legend) at (axis description cs:0.97,0.006);
		\addplot[only marks,
		mark=otimes*, violet,
		mark size=3.5pt
		]
		coordinates {
			(83,43.3)};\label{plot:nae}
		\addplot[only marks,
		mark=otimes*, pink,
		mark size=3.5pt
		]
		coordinates {
			(98,44.0)};\label{plot:nae+}
		\addplot[only marks,
		mark=otimes*, blue,
		mark size=3.5pt
		]
		coordinates {
			(61,45.9)};\label{plot:alignps}
		\addplot[only marks,
		mark=otimes*, cyan,
		mark size=3.5pt
		]
		coordinates {
			(66,46.9)};\label{plot:dmrn} 
		\addplot[only marks,
		mark=otimes*, purple,
		mark size=3.5pt
		]
		coordinates {
			(86,46.7)};\label{plot:seqnet} 	
		\addplot[only marks,
		mark=otimes*, orange,
		mark size=3.5pt
		]
		coordinates {
			(56,49.5)};\label{plot:pstr}
		\addplot[only marks,
		mark=otimes*, green,
		mark size=3.5pt
		]
		coordinates {
			(90,53.3)};\label{plot:coat}
		\addplot[only marks,
		mark=triangle*, red,
		mark size=6pt
		]
		coordinates {
			(48,55.0)};\label{plot:LEAPS}
		\end{axis}
		\node[draw=none,fill=none,anchor= south east] at (legend){\resizebox{5.0cm}{!}{ \begin{tabular}{l|c|c}
				Method  & mAP & Time \\ \hline
				\ref{plot:nae} NAE~\cite{Chen_NAE_CVPR_2020}  & 43.3 & 83   \\
				\ref{plot:nae+} NAE+~\cite{Chen_NAE_CVPR_2020}   & 44.0 & 98   \\
				\ref{plot:alignps} AlignPS~\cite{Yan_AlignPS_CVPR_2021}   & 45.9 & 61   \\
				\ref{plot:dmrn} DMRN~\cite{Han_DMRN_AAAI_2021}   & 46.9 & 66   \\
				\ref{plot:seqnet}
				SeqNet~\cite{Li_SeqNet_AAAI_2021}   & 46.7 & 86  \\
				\ref{plot:pstr} PSTR \cite{Cao_PSTR_CVPR_2022}  & 49.5 & 56   \\ 
                \ref{plot:coat} COAT \cite{Yu_COAT_CVPR_2022} & 53.3 & 90   \\ 
				\hline
				\ref{plot:LEAPS} \textbf{LEAPS} (ours)  & \textbf{55.0} & \textbf{48}   \\ 
				\end{tabular}  }};
		\end{tikzpicture}  }
\end{minipage}
\begin{minipage}{.475\linewidth}
	\resizebox{4.2cm}{!}{%
		\begin{tikzpicture} 
		\begin{axis}[
		axis lines = left,
		ymin=60, ymax=90, 
		xmin=45, xmax=105,
		xlabel=Inference time (ms),
		ylabel= Top-1 Accuracy,
		]
		\addplot[only marks,
		mark=otimes*, violet,
		mark size=3.5pt
		]
		coordinates {
			(83,80.9)};\label{plot:nae1}
		\addplot[only marks,
		mark=otimes*, pink,
		mark size=3.5pt
		]
		coordinates {
			(98,81.1)};\label{plot:nae+1}
		\addplot[only marks,
		mark=otimes*, blue,
		mark size=3.5pt
		]
		coordinates {
			(61,81.9)};\label{plot:alignps1}
		\addplot[only marks,
		mark=otimes*, cyan,
		mark size=3.5pt
		]
		coordinates {
			(66,83.3)};\label{plot:dmrn1} 
		\addplot[only marks,
		mark=otimes*, purple,
		mark size=3.5pt
		]
		coordinates {
			(86,83.4)};\label{plot:seqnet1} 			 
		\addplot[only marks,
		mark=otimes*, orange,
		mark size=3.5pt
		]
		coordinates {
			(56,87.8)};\label{plot:pstr1} 			 
		\addplot[only marks,
		mark=otimes*, green,
		mark size=3.5pt
		]
		coordinates {
			(90,87.4)};\label{plot:coat1}
		\addplot[only marks,
		mark=triangle*, red,
		mark size=6pt
		]
		coordinates {
			(48,88.9)};\label{plot:LEAPS1}
		\end{axis}
		\node[draw=none,fill=none,anchor= south east] at (legend){\resizebox{5.0cm}{!}{ \begin{tabular}{l|c|c}
				Method  & Top-1 & Time \\ \hline
				\ref{plot:nae1} NAE~\cite{Chen_NAE_CVPR_2020}  & 80.9 & 83   \\
				\ref{plot:nae+1} NAE+\cite{Chen_NAE_CVPR_2020}   & 81.1 & 98   \\
				\ref{plot:alignps1} AlignPS~\cite{Yan_AlignPS_CVPR_2021}   & 81.9 & 61   \\
				\ref{plot:dmrn1} DMRN~\cite{Han_DMRN_AAAI_2021}   &83.3 & 66   \\
				\ref{plot:seqnet1}
				SeqNet~\cite{Li_SeqNet_AAAI_2021}   & 83.4 & 86  \\
				\ref{plot:pstr1} PSTR \cite{Cao_PSTR_CVPR_2022}  & 87.8 & 56   \\
                \ref{plot:coat1} COAT \cite{Yu_COAT_CVPR_2022}  & 87.4 & 90   \\
				\hline				
                \ref{plot:LEAPS1} \textbf{LEAPS} (ours)   & \textbf{88.9} & \textbf{48}   \\
				\end{tabular}  }};
		\end{tikzpicture}}
	\centering  \footnotesize  \vspace{-0.2cm}
\end{minipage}
}\vspace{-0.3cm}
\caption{Accuracy (AP) vs. speed (ms) comparison with existing one-step methods on PRW test set \cite{Zheng_PRW_CVPR_2017}.  We show accuracy in terms of mAP (left) and top-1 accuracy (right). All methods use the backbone ResNet50 and the speed is reported on a single NVIDIA V100 GPU.  Our LEAPS outperforms these existing one-step methods in terms of both speed and accuracy.} \vspace{-0.3cm}
\label{fig:speed}
\end{figure}
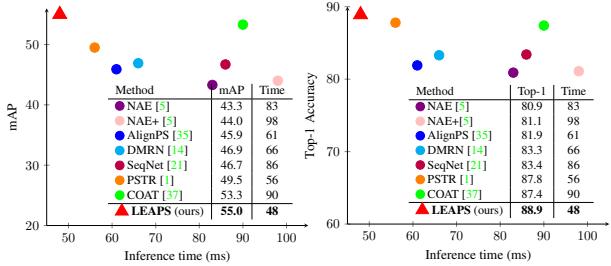

\noindent{\textbf{Contribution:}} We propose an effective yet efficient end-to-end person search framework, named LEAPS. As in  Fig. \ref{fig:intro}(b), Our LEAPS takes  a set of sparse and learned proposals as inputs, and directly predicts person bounding-boxes and their corresponding re-id features with an introduced dynamic person search head. The learnable proposals contain learnable proposal features and learnable proposal boxes, which are randomly initialized at start of training, optimized during training, and finally fixed for inference.
In dynamic person search head, we introduce a novel flexible re-id head to fully exploit discriminative inter-instance and intra-instance information. In addition, instead of directly using one-to-one bipartite matching as detection, we propose a diverse re-id sample matching strategy  that assigns diverse proposals to one ground-truth for re-id feature learning. Compared to the recently end-to-end approach PSTR \cite{Cao_PSTR_CVPR_2022}, our LEAPS are proposed-based method, which fully exploits intra-instance and inter-instance information with a flexible re-id 
head and also introduces a diverse re-id sample matching for better re-id feature learning.  We validate the effectiveness of proposed LEAPS on two standard benchmarks: CUHK-SYSU \cite{Xiao_OIM_CVPR_2017} and PRW \cite{Zheng_PRW_CVPR_2017}. Our LEAPS sets a new state-of-the-art performance on PRW, while operating at a fast speed (see Fig. \ref{fig:speed}). When using the  ResNet50 backbone, our LEAPS has a mAP score of 55.0\% on PRW,  while running at an inference speed of 48 $ms$.  When using the transformer backbone PVTv2-B2, our LEAPS obtains a mAP score of 59.5\% on PRW. 


\section{Related Work}

\noindent\textbf{Person Search:} Most existing methods can be divided into two-step  \cite{Zheng_PRW_CVPR_2017,Cheng_TCTS_CVPR_2020} and one-step \cite{Xiao_OIM_CVPR_2017,Yan_AlignPS_CVPR_2021} approaches. Two-step approaches adopt two independent networks for detection and re-id, respectively. Zheng \textit{et al.} \cite{Zheng_PRW_CVPR_2017} performed an initial two-step attempt by respectively using a detection network and a re-id network for person search. Chen \textit{et al.} \cite{Chen_MGTS_ECCV_2018} first performed person detection and segmentation, and second employed a segmentation-guided module for person re-id. Dong \textit{et al.} \cite{Dong_IGPN_CVPR_2020} introduced an instance guided proposal network to improve detection network. Generally, two-step approaches avoid task conflict between person detection and re-id, but suffer from relatively complicated operations and are usually time-consuming.

\begin{figure*}[t!]
\centering
\includegraphics[width=1.0\linewidth]{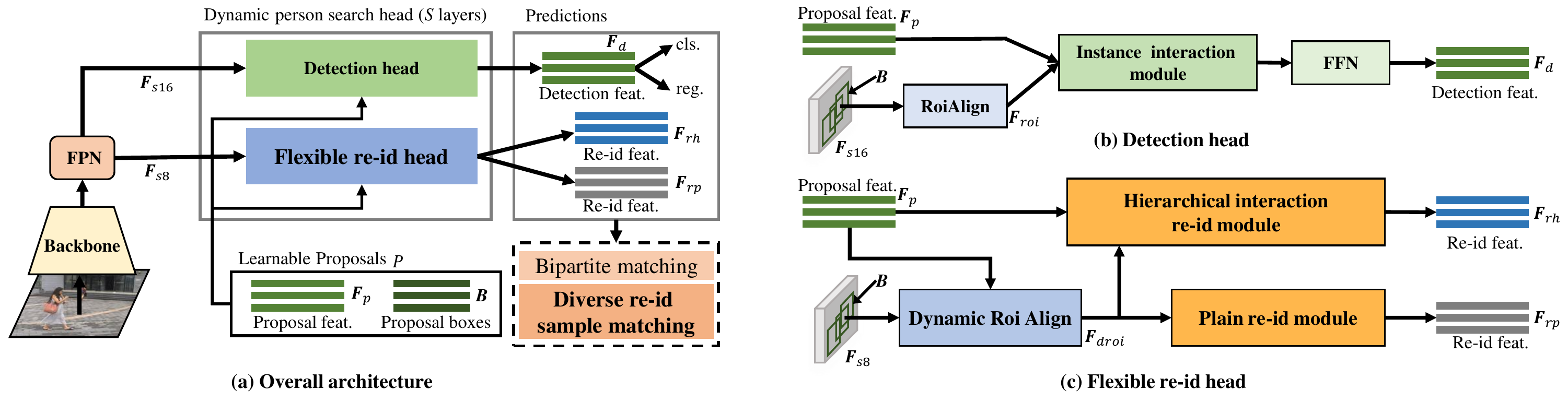} \vspace{-0.6cm}
\caption{(a) Overall architecture of our end-to-end person search framework: LEAPS. It takes deep feature maps $\boldsymbol{F}_{s16},\boldsymbol{F}_{s8}$ and learnable proposals $P$ as inputs, and employs a dynamic person search head for detection and re-id feature generation. The dynamic person search head consists of a  detection head (b) and a flexible re-id head (c). The flexible re-id head first employs a dynamic RoIAlign  to extract discriminative RoI features of the proposals and then predict corresponding re-id features with plain re-id module and hierarchical interaction re-id module. In addition, we introduce a diverse re-id sample matching strategy for re-id feature learning.}
\label{fig:arch}
\end{figure*}

Compared to two-step approaches, one-step approaches are simple, which aim to unify person detection and re-id into a single network. Most one-step approaches are proposal-based methods, which are built on two-stage Faster R-CNN framework \cite{Ren_FasterRCNN_NIPS_2015}. For instance, Xiao \textit{et al.} \cite{Xiao_OIM_CVPR_2017} earlier explored to jointly optimize detection and re-id  in a single network by adding a re-id branch after RoI head. Chen \textit{et al.} \cite{Chen_NAE_CVPR_2020} introduced a norm-aware embedding to decompose detection branch and re-id branch. Li and Miao \cite{Li_SeqNet_AAAI_2021} proposed a sequential headnetwork by first performing person detection and then conducting person re-id. Yu \textit{et al.} \cite{Yu_COAT_CVPR_2022} employed the transformers \cite{Dosovitskiy_ViT_ICLR_2020} to learn re-id features from discriminative person parts in person RoI region.
These proposal-based methods need to design default anchors. To address this issue, Yan \textit{et al.} \cite{Yan_AlignPS_CVPR_2021} extended anchor-free detector FCOS \cite{Tian_FCOS_ICCV_2019} for person search based on a re-id first design principle. These methods still need the hand-designed components, such as NMS post-processing. To avoid NMS, Cao \textit{et al.} \cite{Cao_PSTR_CVPR_2022} performed a first attempt on end-to-end person search with shared transformer re-id decoder structure, named PSTR.
However, PSTR still lags behind the proposal-based one-step methods AGWF \cite{Han_AGWF_ICCV_2021} and COAT \cite{Yu_COAT_CVPR_2022} on PRW dataset. Compared to PSTR, our LEAPS fully exploits intra- and inter-instance information  and performs a diverse re-id sample matching, which significantly outperforms PSTR.
In addition, few works \cite{Yan_CGPS_AAAI_2022,Han_RSiamNets_ICCV_2021,Li_DAPS_ECCV_2022}  explored weakly supervised or domain-aware person search.

\noindent\textbf{End-to-End Object Detection:} Object detection has achieved great success in past years. Early object detection methods \cite{Ren_FasterRCNN_NIPS_2015,Tian_FCOS_ICCV_2019} need the hand-designed components, such as NMS operation or  default anchors. Recently,  end-to-end object detection approaches have been proposed without these hand-designed components. Carion \textit{et al.} \cite{Carion_DETR_ECCV_2020} proposed to treat object detection as a set prediction problem and employed a transformer encoder-decoder to directly output the final set of predictions, called DETR. Zhu \textit{et al.} \cite{Zhu_DeformableDETR_ICLR_2021} replaced the self-attention in DETR with deformable attention to provide better performance, especially on small objects, with less training epochs. Sun \textit{et al.} \cite{Sun_Sparse_CVPR_2021} employed learned object proposals to directly predict objects without NMS operation and object candidates design.

\section{Method}

\noindent \textbf{Overall architecture:}
Fig. \ref{fig:arch}(a) presents the overall architecture of our  end-to-end person search framework with learnable proposals, named LEAPS. We extend the end-to-end object detector Sparse R-CNN \cite{Sun_Sparse_CVPR_2021} for person search by replacing the original detection head with the introduced dynamic person search head. Given an input image $\boldsymbol{I} \in \mathbb{R}^{3\times H\times W}$, we first take a deep backbone, such as ResNet50 \cite{He_ResNet_CVPR_2016}, along with feature pyramid structure \cite{Lin_FPN_CVPR_2017} to extract deep feature maps $\boldsymbol{F}_{s16}\in \mathbb{R}^{256\times H/16 \times W/16},\boldsymbol{F}_{s8}\in \mathbb{R}^{256\times H/8 \times W/8}$. The dynamic person search head takes  deep feature maps $\boldsymbol{F}_{s16},\boldsymbol{F}_{s8}$ and $N$ learnable proposals $P$ as inputs, and predict classification scores, bounding-box regressions, and two diverse re-id features $\boldsymbol{F}_{rh}\in \mathbb{R}^{N\times 256},\boldsymbol{F}_{rp}\in \mathbb{R}^{N\times 256}$.  The dynamic person search head consists of a  detection head and a novel flexible re-id head. The learnable proposals $P$ contains proposal features $\boldsymbol{F}_p\in \mathbb{R}^{N\times 256}$ and proposal boxes $\boldsymbol{B}\in \mathbb{R}^{N\times 4}$, which are randomly initialized at start of training and optimized during training.  After
training,  $\boldsymbol{F}_p$ and $\boldsymbol{B}$ are fixed for inference. Similar to Sparse R-CNN \cite{Sun_Sparse_CVPR_2021}, we adopt cascade structure to stack $S$ dynamic person search heads. In cascade structure, the newly generated detection features $\boldsymbol{F}_{d}$ and boxes at current stage are served as proposal features $\boldsymbol{F}_{p}$ and proposal boxes $\boldsymbol{B}$ at next stage. During inference, we only use the predictions
at last dynamic person search head, where two diverse re-id features are concatenated together for final re-id matching. In addition, we introduce a diverse re-id sample matching strategy to guide  re-id feature learning, instead of directly using bipartite matching as detection.

\subsection{Dynamic Person Search Head}
Dynamic person search head contains a  detection head and a novel flexible re-id head. We adopt two different feature maps for detection and re-id to reduce task conflict. Specifically, the low-resolution feature map $\boldsymbol{F}_{s16}$ is used for detection, and high-resolution feature map $\boldsymbol{F}_{s8}$ is employed for fine re-id. With $\boldsymbol{F}_{s16}$ and learnable proposals $P$, the detection head generates detection features $\boldsymbol{F}_{d}$ and performs classification and regression. With $\boldsymbol{F}_{s8}$ and learnable proposals $P$, the novel flexible re-id head generates discriminative re-id features.

\subsubsection{Detection Head}
The  detection head is similar to the detection head in Sparse R-CNN \cite{Sun_Sparse_CVPR_2021}. As shown in Fig. \ref{fig:arch}(b), the detection head takes feature map $\boldsymbol{F}_{s16}$, proposal features $\boldsymbol{F}_p$, and proposal boxes $\boldsymbol{B}$ as the inputs. With feature map $\boldsymbol{F}_{s16}$ and proposal boxes $\boldsymbol{B}$, we employ RoIAlign operation to extract RoI features $\boldsymbol{F}_{roi}$. Then, an instance  interaction module, followed by a FFN layer, is used to generate output features $\boldsymbol{F}_d$. The instance interaction module consists of two dynamic $1 \times 1$ convolutions with ReLU activation function and a fully-connected layer. The parameters of these two convolutions are dynamically generated by proposal features $\boldsymbol{F}_p$.   The FFN layer contains two fully-connected  (FC) layers with ReLU activation function, where the weights in FC layers are initialized with fixed size.  Finally, the output features $\boldsymbol{F}_d$ are used for proposal classification and regression. In addition, the output features $\boldsymbol{F}_d$ of current stage is used as the proposal feature $\boldsymbol{F}_p$ of next stage in a cascade structure.

\subsubsection{Flexible Re-id Head}\label{sec:FRH}
To encode rich person instance information, we introduce a novel flexible re-id head for diverse re-id feature generation. As shown in Fig. \ref{fig:arch}(c), our flexible re-id head employs dynamic RoIAlign operation to extract discriminative RoI features $\boldsymbol{F}_{droi}$ of proposals. Afterwards, the RoI features $\boldsymbol{F}_{droi}$ are fed to two different branches for re-id feature generation. One branch directly uses a plain re-id module to extract the re-id features $\boldsymbol{F}_{rp}$, whereas the other branch adopts a hierarchical interaction re-id module to generate the re-id features $\boldsymbol{F}_{rh}$. These two branches can exploit rich instance information. To learn diverse features, we employ a diverse re-id loss during training. During inference, we concatenate these two features as final re-id features.

\begin{figure}[t!]
\centering
\includegraphics[width=1.0\linewidth]{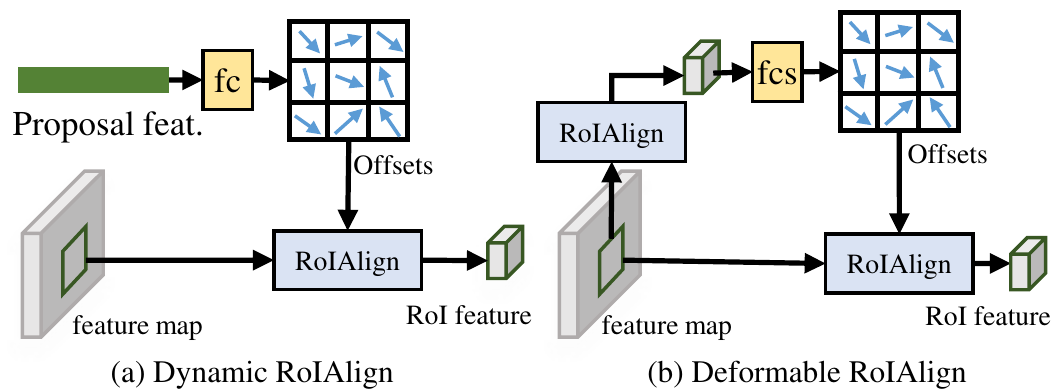} \vspace{-0.6cm}
\caption{(a) Our dynamic RoIAlign operation and (b) deformable RoIAlign operation \cite{Dai_DCN_ICCV_2017}. Compared to deformable RoIAlign, our dynamic RoIAlign directly predicts the bin offsets using proposal features, which does not requires extra RoIAlign operation for bin offset prediction.}
\label{fig:roi}
\end{figure}

\noindent\textbf{Dynamic RoIAlign:} Most existing one-step person search approaches are built on Faster R-CNN framework \cite{Ren_FasterRCNN_NIPS_2015,Lin_FPN_CVPR_2017} and adopt standard RoI pooling or RoIAlign operation to extract RoI features of proposals. The RoI pooling or RoIAlign operation divides the proposals into  fixed-grid bins, which struggles to effectively deal with person  deformations. To address this issue, we introduce a dynamic RoIAlign operation to dynamically extract RoI features according to proposal features. We first use proposal features $\boldsymbol{F}_p$ to predict  bin offsets $\Delta \boldsymbol{p}$. Afterwards we employ RoIAlign  to extract RoI features from the offset bins. The RoI feature $\boldsymbol{F}_{droi}^n$ of proposal $n$ can be written as
\begin{equation}
    \boldsymbol{F}_{droi}^n(i,j) = \text{RoIAlign}(\boldsymbol{F}_{s8},\boldsymbol{p}^n_{ij}+\Delta \boldsymbol{p}^n_{ij}),
\end{equation}
where $\boldsymbol{p}$ represents the original locations of  bins, $i,j$ is the index of the bin, and  $\Delta \boldsymbol{p}$ represents $x$-axis and $y$-axis offsets of each bin. Fig. \ref{fig:roi} compares our dynamic RoIAlign with deformable RoIAlign \cite{Dai_DCN_ICCV_2017}. Different to deformable RoIAlign, our dynamic RoIAlign  does not need a separate RoIAlign operation to predict  bin offsets. As a result, our dynamic RoIAlign is  more efficient. We observe that our dynamic RoIAlign has a similar performance with deformable RoIAlign with 2$\times$ speedup.

\noindent\textbf{Plain re-id  module:} With the RoI features $\boldsymbol{F}_{droi}$, a  plain re-id  module (PRM) to predict re-id feature is using two-stacked fully-connected layer to generate the plain re-id features $\boldsymbol{F}_{rp}$. However, we argue that this plain re-id feature module does not fully exploit inter-instance and intra-instance information. To this end, we introduce a novel hierarchical interaction re-id feature module.

\begin{figure}[t]
\centering
\includegraphics[width=1.0\linewidth]{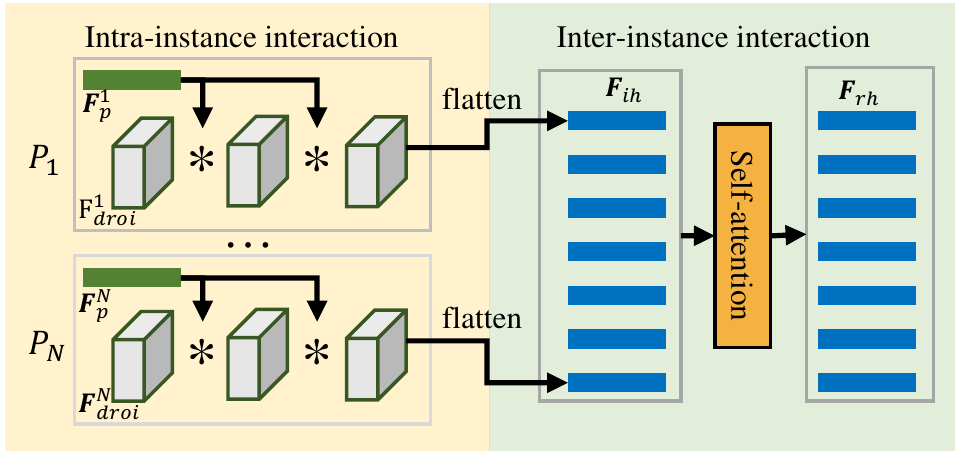} \vspace{-0.6cm}
\caption{Detailed structure of our hierarchical interaction re-id module (HIRM). HIRM first performs intra-instance interaction and then performs inter-instance interaction. As a result, it fully exploits inter-instance and intra-instance information for re-id.} \vspace{-0.4cm}
\label{fig:hirm}
\end{figure}

\noindent\textbf{Hierarchical interaction re-id  module:} To better encode instance information for re-id, we introduce a hierarchical interaction re-id  module  (HIRM) that performs both inter-instance and intra-instance information interactions. Fig. \ref{fig:hirm} presents the detailed structure of the proposed HIRM. With proposal features $\boldsymbol{F}_p$ and RoI features $\boldsymbol{F}_{droi}$, we first perform intra-instance information interaction using two dynamic $1\times1$ convolutions. The parameters of two dynamic convolutions are generated by proposal features $\boldsymbol{F}_p$ with a fully-connected layer. As a result, different proposals have different parameters to perform intra-instance interaction in each proposal. With intra-instance information interaction, we generate the intermediate features $\boldsymbol{F}_{ih}$ and flatten the features $\boldsymbol{F}_{ih}$ along the dimensions of width, height, and channel. Then, we employ a self-attention module to perform inter-instance information interaction. The self-attention module captures the relationship between different inputs, thereby exploiting the context information between different instances. The output features after inter-instance interaction are represented as the re-id features $\boldsymbol{F}_{rh}$.

The flexible re-id head predicts two different re-id features $\boldsymbol{F}_{rp}$ and $\boldsymbol{F}_{rh}$.  During inference, we concatenate these two features together for re-id matching. When these two features have complementary characteristics, the concatenated features are likely to perform more accurate re-id matching. To guide diverse  re-id feature learning, we employ a diverse re-id loss \cite{Li_Diverse_CVPR_2021} that minimizes the similarity between the intermediate features in PRM and HIRM as $L_{div}=\text{cos}(\boldsymbol{F}_{ip},\boldsymbol{F}_{ih}^{n})$, where $\boldsymbol{F}_{ip}$ is the  feature after first fully-connected layer in PRM, $\boldsymbol{F}_{ih}$ is the intermediate feature in HIRM.  The cosine similarity is inversely proportional to feature difference (\eg, cosine similarity equals to zero implies that two features are orthogonal, and vice versa). Therefore, minimizing  cosine similarity  can likely lead to  diverse re-id features.

\begin{figure}[t!]
\centering
\begin{subfigure}{0.49\linewidth}
\includegraphics[width=\linewidth]{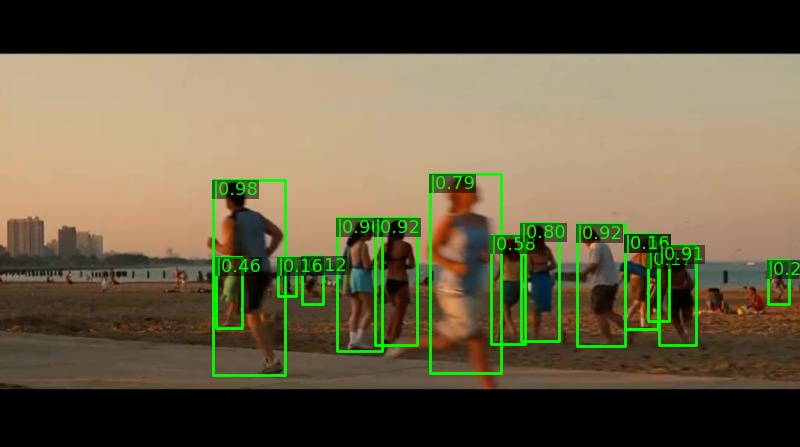}
\caption{$0.1$}
\end{subfigure}
\begin{subfigure}{0.49\linewidth}
\includegraphics[width=\linewidth]{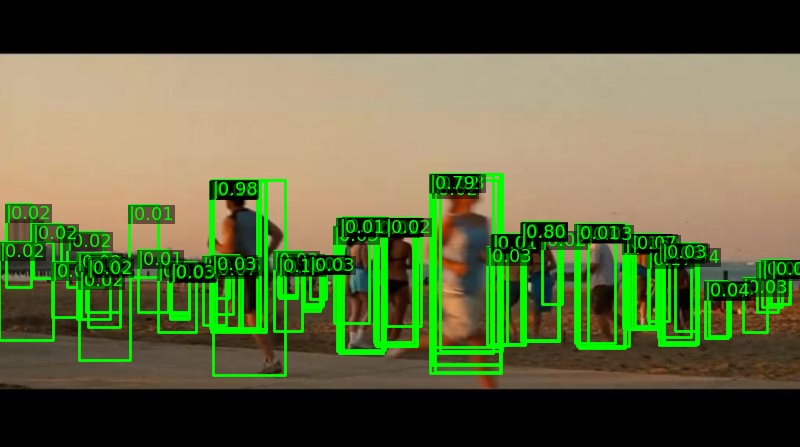}
\caption{$0.01$}
\end{subfigure}
\vspace{-0.3cm}
\caption{Detection results of end-to-end object detector. We show the bounding-boxes with the classification scores larger than the given thresholds (\textit{i.e.,}  0.1, 0.01). In end-to-end object detector, except one bounding-box with high score, there are some predicted bounding-boxes with low scores around the persons. We argue that these bounding-boxes can provide diverse instance information for re-id.} \vspace{-0.4cm}
\label{fig:diver}
\end{figure}

\subsection{Diverse Re-id Sample Matching}\label{sec:DRSM}

We introduce a diverse re-id sample matching (DRSM) strategy to better guide re-id feature learning in our end-to-end LEAPS. The diverse sample matching strategy aims to incorporate diverse samples of each person instance into the training. Generally, the end-to-end object detectors \cite{Sun_Sparse_CVPR_2021,Carion_DETR_ECCV_2020}  adopt one-to-one bipartite matching between predictions and ground-truths. A straightforward way is to directly adopt one-to-one sample matching for both detection and re-id. As a result, only one positive sample per person instance is used for re-id feature learning.  We argue that,  compared to detection that separates persons from the background, re-id needs more rich instance information to distinguish different person instances. Therefore,  one  positive sample is not sufficient to guide discriminative re-id feature learning. To address this issue, we adopt  one-to-one bipartite matching for detection and introduce diverse re-id sample matching for re-id head.

Fig. \ref{fig:diver} shows example detection results from end-to-end object detector \cite{Sun_Sparse_CVPR_2021} under different classification thresholds. With very low threshold (\textit{e.g.,} 0.01), we observe that there exists multiple different proposals around each person instance. Here, we argue that these bounding-boxes with low scores can be used for re-id feature learning. Instead of one-to-one sample matching as in detection head, our diverse re-id sample matching strategy assigns the proposal with the identity label of corresponding ground-truth when the proposal has an IoU overlap higher than $\theta=0.7$ with the ground-truth. As a result, each person corresponds to multiple proposals, which  provide diverse instance information for re-id feature learning.
Instead of treating each assigned proposal equally for re-id learning, we argue that the proposals with more accurate localization should contribute more. Therefore, we further design a  weighting strategy to pay more attention on these samples with accurate localization, where the re-id weight of proposal $P^n$ is $W^n=(IoU^{n}-\theta)/(1-\theta)+0.5$. Namely, the proposal with higher IoU overlap has a larger weight. The proposed diverse re-id sample matching marginally increases complexity during training and does not affect inference speed.

\subsection{Training and Inference}

During training, we first perform an
optimal one-to-one bipartite matching between proposals and ground-truths as in Sparse R-CNN \cite{Sun_Sparse_CVPR_2021}. As a result, one person only corresponds to one positive proposal for detection and re-id feature learning.  Afterwards, we perform diverse re-id sample matching to assign each person with more positive proposals for re-id sub-task. Similar to Sparse R-CNN \cite{Sun_Sparse_CVPR_2021}, we employ the same losses for person detection, which contain focal loss for classification, L1 loss and
generalized IoU loss for regression.  To guide re-id feature learning, we build a lookup table $V\in\mathbb{R}^{L\times 256}$ and a circular queue $U\in\mathbb{R}^{Q\times 256}$. The lookup table $V$ stores re-id features of $L$ labeled person identities, while the circular queue $U$ stores re-id features of $Q$ unlabelled person identities. Similar to \cite{Xiao_OIM_CVPR_2017}, lookup table $V$ and circular queue $U$ are updated at each iteration.  Based on lookup table $V$ and circular queue $U$, we use the triplet-aided
online instance matching (OIM) loss \cite{Yan_AlignPS_CVPR_2021} for re-id feature learning. The overall losses are written as
\begin{equation}
    L=\sum_{s} L_{det}^{s}+\lambda [s>4](WL_{toim}^s+L_{div}^s),
\end{equation}
where $L_{det}$ is detection loss, $L_{toim}$ is triplet-aided OIM loss \cite{Yan_AlignPS_CVPR_2021} for re-id, $W$ is the re-id weight in Sec. \ref{sec:DRSM}, and $s$ represents the stage index. The bracket indicator function
$[s > 4]$  is equal to 1 when $s > 4$, and 0 otherwise.

During inference, person search aims to search a box-annotated target person in a query image from  gallery images. We first employ LEAPS to predict person bounding-boxes and their corresponding re-id features. Afterwards, we assign the predicted re-id feature to the target person if the corresponding bounding-box has the maximum overlap with the target person. Similarly, we can generate the re-id features of persons in gallery images. Finally, we match the target person in query image with the persons in gallery images according to cosine similarities.

\begin{table}[t!]
\renewcommand{\arraystretch}{1.0}
\begin{center}
\resizebox{\linewidth}{!}{
\begin{tabular}{|l|c|cc|cc|}
\hline
\multirow{2}*{Method}       & \multirow{2}*{Backbone}      & \multicolumn{2}{c|}{CUHK-SYSU} & \multicolumn{2}{c|}{PRW}  \\\cline{3-6}
       &       & mAP   & Top-1 & mAP & Top-1  \\
\hline \hline
\multicolumn{6}{|l|}{\textit{Two-step}}\\
IDE \cite{Xiao_OIM_CVPR_2017}  & ResNet50    &  - & - & 20.5 & 48.3 \\ 
MGTS \cite{Chen_MGTS_ECCV_2018}   & VGG16    &  83.0 & 83.7 & 32.6 & 72.1 \\ 
CLSA \cite{Lan_CLSA_ECCV_2018}  & ResNet50    &  87.2 & 88.5 & 38.7 & 65.0 \\ 
RDLR \cite{Han_RDLR_ICCV_2019}  & ResNet50    &  93.0 & 94.2 & 42.9 & 70.2 \\ 
IGPN  \cite{Dong_IGPN_CVPR_2020} & ResNet50    &  90.3 & 91.4 & {47.2} & 87.0 \\ 
TCTS \cite{Cheng_TCTS_CVPR_2020}  & ResNet50    &  {93.9} & {95.1} & 46.8 & {87.5} \\ 
\hline \hline
\multicolumn{6}{|l|}{\textit{One-step with NMS}}  \\ 
OIM \cite{Xiao_OIM_CVPR_2017}      & ResNet50    & 75.5 & 78.7  & 21.3 & 49.4 \\
IAN \cite{Xiao_IAN_PR_2019}      & ResNet50    & 76.3 & 80.1  & 23.0 & 61.9 \\
NPSM \cite{Liu_NPSM_ICCV_2017}   & ResNet50    & 77.9 & 81.2  & 24.2 & 53.1 \\
RCAA \cite{Chang_RCAA_ECCV_2018}   & ResNet50    & 79.3 & 81.3  & - & - \\
CTXG  \cite{Yan_CTXG_CVPR_2019}     & ResNet50    & 84.1 & 86.5  & 33.4 & 73.6 \\
QEEPS \cite{Munjal_QEEPS_CVPR_2019}      & ResNet50    & 88.9 & 89.1  & 37.1 & 76.7 \\
HOIM \cite{Wang_HOIM_CVPR_2020}   & ResNet50    & 89.7 & 90.8  &  39.8 & 80.4 \\
APNet \cite{Zhong_APNet_CVPR_2020}      & ResNet50    & 88.9 & 89.3  & 41.9 & 81.4 \\
BINet \cite{Dong_BINet_CVPR_2020}      & ResNet50    & 90.0 & 90.7  & 45.3 & 81.7 \\
NAE \cite{Chen_NAE_CVPR_2020}      & ResNet50    & 91.5 & 92.4  & 43.3 & 80.9 \\
DMRNet \cite{Han_DMRN_AAAI_2021}       & ResNet50    & 93.2 & 94.2  & 46.9 & 83.3 \\
PGSFL \cite{Kim_PGSFL_CVPR_2021}   & ResNet50    & 90.2 & 91.8  & 42.5 & 83.5 \\
SeqNet \cite{Li_SeqNet_AAAI_2021}      & ResNet50    & 93.8 & 94.6  & 46.7 & 83.4 \\
DMRN    \cite{Han_DMRN_AAAI_2021}   & ResNet50    & 93.2 & 94.2  & 46.9 & 83.3 \\
DKD    \cite{Zhang_DKD_AAAI_2021}  & ResNet50    & 93.1 & 94.2  & 50.5 & 87.1 \\
OIMNet++ \cite{Sang_OIM++_ECCV_2022}       & ResNet50    & 93.1 & 93.9  & 46.8 & 83.9 \\
AlignPS \cite{Yan_AlignPS_CVPR_2021}       & ResNet50    & 93.1 & 93.4  & 45.9 & 81.9 \\
AGWF    \cite{Han_AGWF_ICCV_2021}   & ResNet50    & 93.3 & 94.2  & 53.3 & 87.7 \\
COAT    \cite{Yu_COAT_CVPR_2022}   & ResNet50    & 94.2 & 94.7  & 53.3 & 87.4 \\
\hline 
\multicolumn{6}{|l|}{\textit{One-step without NMS}} \\ 
PSTR    \cite{Cao_PSTR_CVPR_2022}   & ResNet50    & 93.5 & 95.0  & 49.5 & 87.8 \\
\textbf{LEAPS} (Ours)       & ResNet50    & \textbf{95.2} & \textbf{95.8}  & \textbf{55.0} & \textbf{88.9} \\
PSTR   \cite{Cao_PSTR_CVPR_2022}  & PVTv2-B2    & 95.2 & 96.2  & 56.5 & \textbf{89.7} \\
\textbf{LEAPS} (Ours)      & PVTv2-B2    & \textbf{96.4} & \textbf{96.9}  & \textbf{59.5} & \textbf{89.7} \\
\hline
\hline
\multicolumn{6}{|l|}{\textit{Cross-view evaluation on PRW}} \\ 

HOIM \cite{Wang_HOIM_CVPR_2020}   & ResNet50    & - & -  &  36.5 & 65.0 \\
NAE \cite{Chen_NAE_CVPR_2020}  & ResNet50    & - & -  &  40.0 & 67.5\\
SeqNet    \cite{Li_SeqNet_AAAI_2021}   & ResNet50    & - & -  & 43.6 & 68.5 \\
AGWF    \cite{Han_AGWF_ICCV_2021}   & ResNet50    & - & -  & 48.0 & 73.2 \\
COAT    \cite{Yu_COAT_CVPR_2022}   & ResNet50    & - & -  & 50.9 & 75.1\\
\textbf{LEAPS} (Ours)   & ResNet50    & - & -  & \textbf{52.6} & \textbf{77.1} \\
\hline
\end{tabular}}\vspace{-0.5cm}
\end{center}
\caption{State-of-the-art comparison in terms of mAP and top-1 accuracy on CUHK-SYSU and PRW test sets. With   ResNet50 backbone, our LEAPS significantly outperforms existing methods, including proposal-based COAT \cite{Yu_COAT_CVPR_2022} and end-to-end PSTR \cite{Cao_PSTR_CVPR_2022}. With PVTv2-B2 backbone, our LEAPS sets a favorable performance on both datasets. We also present the results of cross-view evaluation on PRW.} \vspace{-0.4cm}
\label{tab:sota}
\end{table}

\section{Experiments}
\subsection{Datasets and Implementation Details}
\noindent\textbf{CUHK-SYSU} \cite{Xiao_OIM_CVPR_2017} is a large-scale person search dataset collected from street  and movie snaps, which contains rich variations of viewpoints, lighting, and backgrounds.  The dataset is divided into the training set and test set, which do not have overlapped images or identities. The training set has 11,206 images, 55,272 persons, and 5,532 different identities. The test set contains 6,978 images, 40,871 persons, and 2,900 identities. For performance evaluation, the dataset provides multiple different gallery sizes ranging from 50 to 4,000. We perform detailed experiments with the standard gallery size of 100 following most person search approaches \cite{Xiao_OIM_CVPR_2017,Yan_AlignPS_CVPR_2021,Cao_PSTR_CVPR_2022}. In addition, we present performance comparison under different gallery sizes.

\noindent\textbf{PRW} \cite{Zheng_PRW_CVPR_2017} is a challenging person search dataset collected in Tsinghua university. Six different static cameras are used to capture the images of different views.  The training set contains 5,704 images, 18,048 persons, and 482 identities. The test set has 6,112 images, 25,062 persons, and 450 identities. The dataset provides the query set and corresponding gallery images.

\noindent\textbf{Evaluation metrics:} There are two standard metrics for performance evaluation, including mean Averaged Precision (mAP) and  top-1 accuracy.

\noindent\textbf{Implementation details:} We implement our LEAPS based on the open-source mmdetection \cite{mmdetection}. Two deep models pre-trained on ImageNet \cite{Russakovsky_ImageNet_IJCV_2015} are used as the backbones, including ResNet50 \cite{He_ResNet_CVPR_2016} and PVTv2-B2 \cite{Wang_PVTv2_arXiv_2021}. The optimizer is AdamW with weight decay 0.0001. The model is trained on a single NVIDIA V100 with the mini-batch of 2 images. There are totally 36 epochs and the initial learning rate is set to 0.00003 and decreased by a factor of 10 at epoch 24 and 33. Similar to \cite{Yan_AlignPS_CVPR_2021,Cao_PSTR_CVPR_2022},
we adopt the multi-scale training strategy during the training, and we  rescale the test images to a fixed size of  $1500\times900$ pixels during inference. The number $N$ of learnable proposals are set as 100.

\begin{figure}[t!]
\centering
\includegraphics[width=1.0\linewidth]{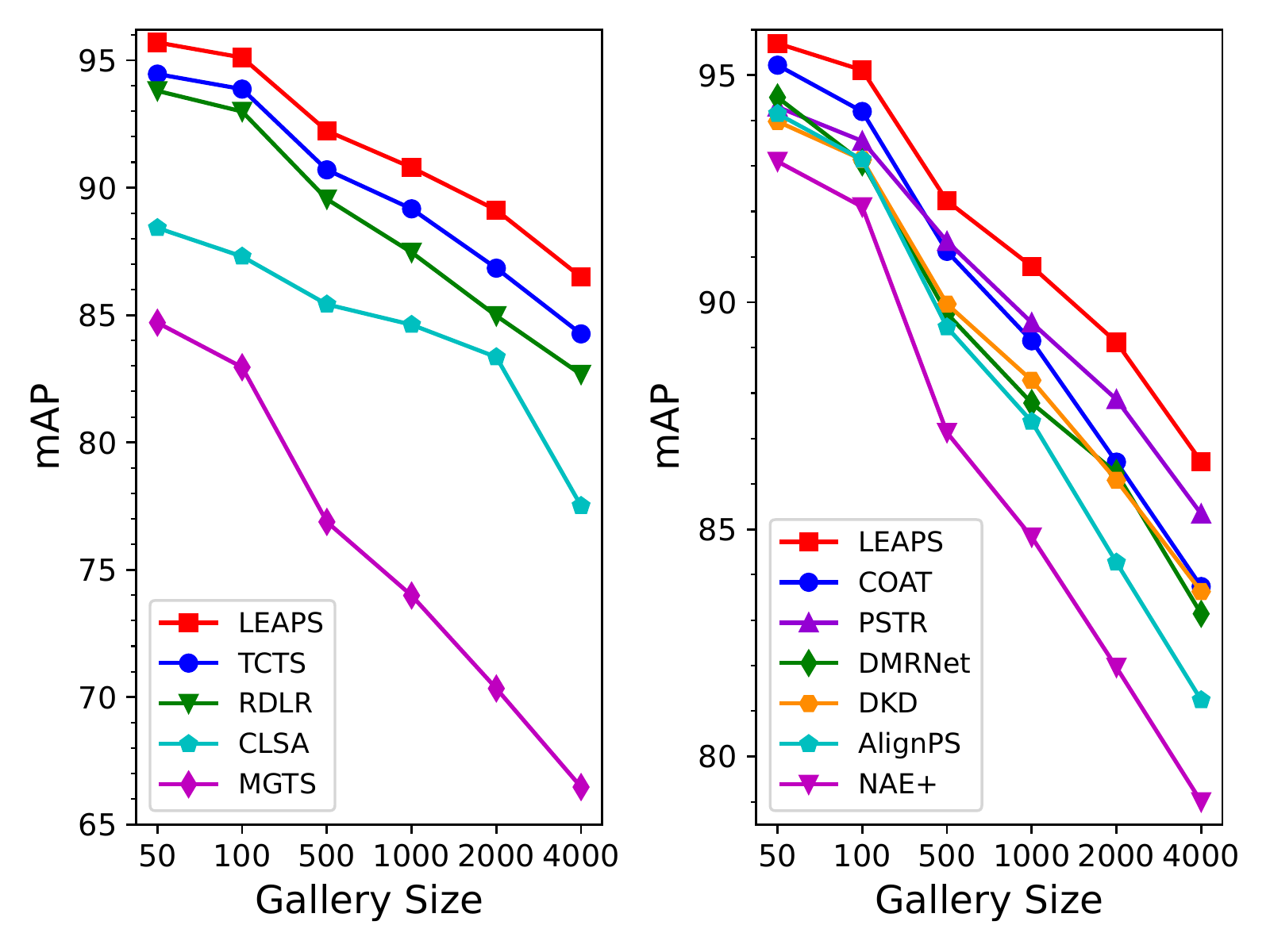} \vspace{-0.6cm}
\caption{Comparison with two-step (left) and one-step (right) approaches on CUHK-SYSU test set under different gallery sizes. All the methods adopt ResNet50 as the backbone (except MGTS which adopts VGG16). Our LEAPS achieves consistent improvement in performance, compared to existing methods.} \vspace{-0.4cm}
\label{fig:curve}
\end{figure}
\begin{table}[t!]
\footnotesize
\renewcommand{\arraystretch}{1.0}
\begin{center}
\begin{tabular}{|c|cc|cc|}
\hline
Number & DRSM  (Sec. \ref{sec:DRSM})    & FRH (Sec. \ref{sec:FRH})           & mAP & Top-1  \\
\hline \hline
(a) &       &       & 43.5 & 81.7 \\
(b) & \checkmark        &      & 46.8 & 86.3 \\
(c) & \checkmark      & \checkmark        & 55.0 & 88.9\\
\hline
\end{tabular}\vspace{-0.5cm}
\end{center}
\caption{Impact of integrating different modules into our LEAPS, including diverse re-id sample matching (DRSM) in Sec. \ref{sec:DRSM} and flexible re-id head (FRH) in Sec. \ref{sec:FRH}. The baseline in (a) adopts a single PRM with one-to-one bipartite  sample matching  for re-id.}\vspace{-0.4cm}
\label{tab:parts}
\end{table}

\subsection{State-of-the-art Comparison}
Here, we present a state-of-the-art comparison  on both CUHK-SYSU \cite{Xiao_OIM_CVPR_2017} and PRW \cite{Zheng_PRW_CVPR_2017} test sets.

\noindent \textbf{Comparison on CUHK-SYSU:} Table \ref{tab:sota} shows state-of-the-art comparison with both two-step and one-step approaches under the gallery size of 100. We divide one-step approaches into two classes according to NMS post-processing. Among  one-step approaches, using the ResNet50 backbone, the recently introduced anchor-free AlignPS \cite{Yan_AlignPS_CVPR_2021} and proposal-based COAT \cite{Yu_COAT_CVPR_2022} achieve the mAP scores of 93.1\% and 94.2\%, respectively. With the same backbone, our LEAPS achieves a mAP score of 95.2\%. Compared to AlignPS and COAT, our LEAPS obtains 2.1\% and 1.0\% improvements in terms of mAP, and does not require hand-designed NMS post-processing. In addition, COAT has an inference speed of 90 $ms$, while our LEAPS obtains an inference speed of 48 $ms$. Therefore, our LEAPS is 1.9 times faster than COAT. The end-to-end PSTR \cite{Cao_PSTR_CVPR_2022} adopts transformer encoder-decoder for person search and achieves a mAP score of 93.5\% using ResNet50. Compared to end-to-end PSTR, our LEAPS obtains absolute gain of 1.7\%  in terms of mAP. We argue that our flexible re-id head can provide rich instance information for re-id. When using the transformer backbone PVTv2-B2, our LEAPS obtains a mAP score of 96.4\%, outperforming PSTR with an absolute gain of 1.2\%.

In addition, Fig. \ref{fig:curve}  presents a detailed comparison on CUHK-SYSU test set under different gallery sizes. We show two-step approaches at  left, and one-step approaches at  right. The larger gallery size indicates searching the target person in more gallery images, which is more challenging. We observe that our LEAPS outperforms these one-step approaches, especially on larger gallery size, which demonstrates that our LEAPS performs better under more challenging settings.

\noindent \textbf{Comparison on PRW:} Table \ref{tab:sota} also shows the results of these state-of-the-art approaches on PRW test set. With the  ResNet50 backbone, the proposal-based methods AGWF \cite{Han_AGWF_ICCV_2021} and COAT \cite{Yu_COAT_CVPR_2022} both have a mAP score of 53.3\%, and the anchor-free method AlignPS has a mAP score of 45.9\%. Using the same backbone, our LEAPS obtains a mAP score of 55.0\%. Our LEAPS outperforms AGWF, COAT, and AlignPS by 1.7\%, 1.7\%, and 9.1\% in terms of mAP. In terms of top-1 accuracy, our LEAPS outperforms AGWF, COAT, and AlignPS by 2.2\%, 2.5\%, and 7.0\%. It demonstrates that our LEAPS outperforms both proposal-based  and anchor-free methods on this dataset. Further, LEAPS achives significant improvement in performance over end-to-end PSTR \cite{Cao_PSTR_CVPR_2022}, obtaining absolute gains of 5.5\% and 3.0\% in terms of mAP using ResNet50 and PVTv2-B2.

The bottom of  Table \ref{tab:sota} further presents the cross-view evaluation on PRW, where the query image and gallery image are from different cameras. Our LEAPS also obtains significant improvements on both mAP and top-1 accuracy.

\subsection{Ablation Study}
Here, we perform extensive ablation study  on PRW test set, and employ ResNet50 as the backbone.

\begin{table}[t!]
\footnotesize
\renewcommand{\arraystretch}{1.0}
\begin{center}
\begin{tabular}{|c|cccc|cc|}
\hline
Num. & Dyn. RoIAlign    & PRM     & HIRM   & $L_{div}$    & mAP & Top-1  \\
\hline \hline
(a)  &      &   \checkmark    & & & 46.8 & 86.3 \\
(b) & \checkmark      & \checkmark      & &  & 49.6 & 87.3 \\
(c) & \checkmark      &       & \checkmark &  & 50.9 & 87.1 \\
(d) & \checkmark      & \checkmark    & \checkmark  &   & 54.5 & 88.6\\
(e) & \checkmark      & \checkmark    & \checkmark  & \checkmark  & 55.0 & 88.9 \\
\hline
\hline
(f) & \multicolumn{4}{|l|}{FRH with two paralleled PRM branches}   & 51.3 & 88.3 \\
(g) & \multicolumn{4}{|l|}{FRH with two paralleled HIRM branches}   & 52.1 & 87.4 \\
(h) &  \multicolumn{4}{|l|}{FRH w/o intra-instance interaction}         &  52.6 & 87.9 \\
(i) &  \multicolumn{4}{|l|}{FRH w/o inter-instance interaction}         &  53.8 & 88.1 \\
\hline
\end{tabular}\vspace{-0.5cm}
\end{center}
\caption{Impact of different components and designs in our flexible re-id head (FRH). The top part shows the impact of different components, including dynamic RoI, plain re-id module (PRM), hierarchical interaction re-id module (HIRM), and diverse re-id loss $L_{div}$. The bottom part presents the impact of some different designs.}\vspace{-0.3cm}
\label{tab:frh}
\end{table}

\begin{table}[t!]
\footnotesize
\renewcommand{\arraystretch}{1.0}
\begin{center}
\begin{tabular}{|c|l|l|cc|}
\hline
Num.     &\multicolumn{2}{|c|}{With or without DRSM}      & mAP & Top-1  \\
\hline
\hline
\multirow{3}*{(a)} & \multirow{3}*{RoIAlign+PRM}&     & 43.5 & 81.7 \\
& & $\dag$     & 46.2  & 86.0 \\
& & $\ddag$     & 46.8 & 86.7 \\
\hline
\multirow{3}*{(b)} & \multirow{3}*{Dy. RoIAlign+PRM}&     & 37.2 & 67.9 \\
& & $\dag$    & 49.4 & 87.1  \\
& & $\ddag$  & 49.6 & 87.3 \\
\hline
\multirow{3}*{(c)} &\multirow{3}*{Dy. RoIAlign+PRM+HIRM}&    & 43.7  & 76.0  \\
& & $\dag$    & 54.3 & 88.6 \\
& & $\ddag$     & 54.5 & 88.6 \\
\hline
\multirow{3}*{(d)} &\multirow{3}*{Dy. RoIAlign+PRM+HIRM+$L_{div}$}&    & 43.9 & 78.6  \\
& & $\dag$     & 54.5 &  88.7 \\
& & $\ddag$     & 55.0 & 88.9 \\
\hline
\end{tabular}\vspace{-0.5cm}
\end{center}
\caption{Impact of diverse re-id sample matching (DRSM) using different re-id heads. $\dag$ represents DRSM without re-id  weighting strategy, and $\ddag$ represents DRSM with re-id weighting. DRSM provides significant improvements, especially under the dynamic feature extraction in (b-d).} \vspace{-0.4cm}
\label{tab:dsm}
\end{table}

\noindent \textbf{Integrating different modules into our LEAPS:}  Table \ref{tab:parts} shows the results of integrating diverse re-id sample matching (DRSM) and flexible re-id head (FRH) into our LEAPS. In  the baseline (a), we directly add a single PRM after RoI features for re-id and adopt the same one-to-one bipartite sample matching for both detection and re-id. The baseline achieves a mAP score of 43.5\% and top-1 accuracy of 81.7\%. When integrating DRSM into the baseline, it (b) achieves 46.8\% on mAP and 86.3\% on top-1 accuracy, which outperforms the baseline by 3.3\% and 4.6\%, respectively. When further integrating FRH into the baseline, it (c) outperforms the baseline by 11.5\% and 7.2\%.

\noindent \textbf{Impact of flexible re-id head:}  Table \ref{tab:frh} first shows the impact of different components in  flexible re-id head (FRH). When using dynamic RoIAlign to replace RoIAlign for re-id, it (b) has 2.8\% improvement on mAP and 1.0\% improvement on top-1 accuracy. With dynamic RoIAlign and hierarchical interaction re-id module (HIRM), it (c) has 4.1\% improvement on mAP and 0.8\% improvement on top-1 accuracy, respectively. With integrating these two components together, it (d) obtains a mAP score of 54.5\%  and a top-1 accuracy of 88.6\%. It demonstrates that our PRM and HIRM can provide diverse features for re-id. With further incorporating diversity feature loss, it obtains  0.5\%  and 0.3\% improvements on mAP and top-1 accuracy.

In the bottom of Table \ref{tab:frh}, we also present the impact of different designs in our flexible re-id head (FRH).  When we adopt two paralleled PRMs, it (f) achieves a mAP score of  51.3\%  and top-1 accuracy of 88.3\%. When we adopt two paralleled HIRMs, it (g) achieves a mAP score of  52.1\%  and top-1 accuracy of 87.4\%. It can be seen that both (f) and (g) are inferior to (e).  It further demonstrates that two paralleled PRM and HIRM can provide diverse features. In addition, we show the results of removing the intra-instance interaction (h) or  inter-instance interaction (i). We also observe the performance drop on both mAP and top-1 accuracy.

\begin{table}[t!]
\footnotesize
\renewcommand{\arraystretch}{1.0}
\begin{center}
\begin{tabular}{|l|ccccccc|}
\hline
Stage & 2& 3& 4 & 5 & 6 & 7 & 8 \\
\hline \hline
mAP  & 18.5  & 34.2  & 50.4 & 53.9 & 55.0 & 54.9 & 54.8\\
Top-1 & 63.7  & 77.4  & 86.3 & 88.8 & 88.9 & 88.9 & 88.7\\
Time (ms) &  38  & 41 & 44 & 46 & 48 & 51 & 54\\
\hline
\end{tabular}\vspace{-0.5cm}
\end{center}
\caption{Impact of different cascade stages in our LEAPS. We adopt cascade structure with 6 stages.}\vspace{-0.3cm}
\label{tab:cascade}
\end{table}

\begin{table}[t!]
\footnotesize
\renewcommand{\arraystretch}{1.0}
\begin{center}
\begin{tabular}{|l|cccccc|}
\hline
Stage & 1    & 2     & 3      & 4 & 5 & 6  \\
\hline \hline
mAP  &   52.1   & 55.0  &  54.8 & 52.7 & 47.4 & 46.0 \\
Top-1 & 87.2  & 88.9  & 88.5 & 87.7 & 85.5 & 84.5 \\
\hline
\end{tabular}\vspace{-0.5cm}
\end{center}
\caption{Impact of adding flexible re-id head to  last several stages during training. We add the two re-id heads at last 2 stages.}\vspace{-0.5cm}
\label{tab:reidstage}
\end{table}

\begin{figure}[t!]
\centering
\includegraphics[width=1.0\linewidth]{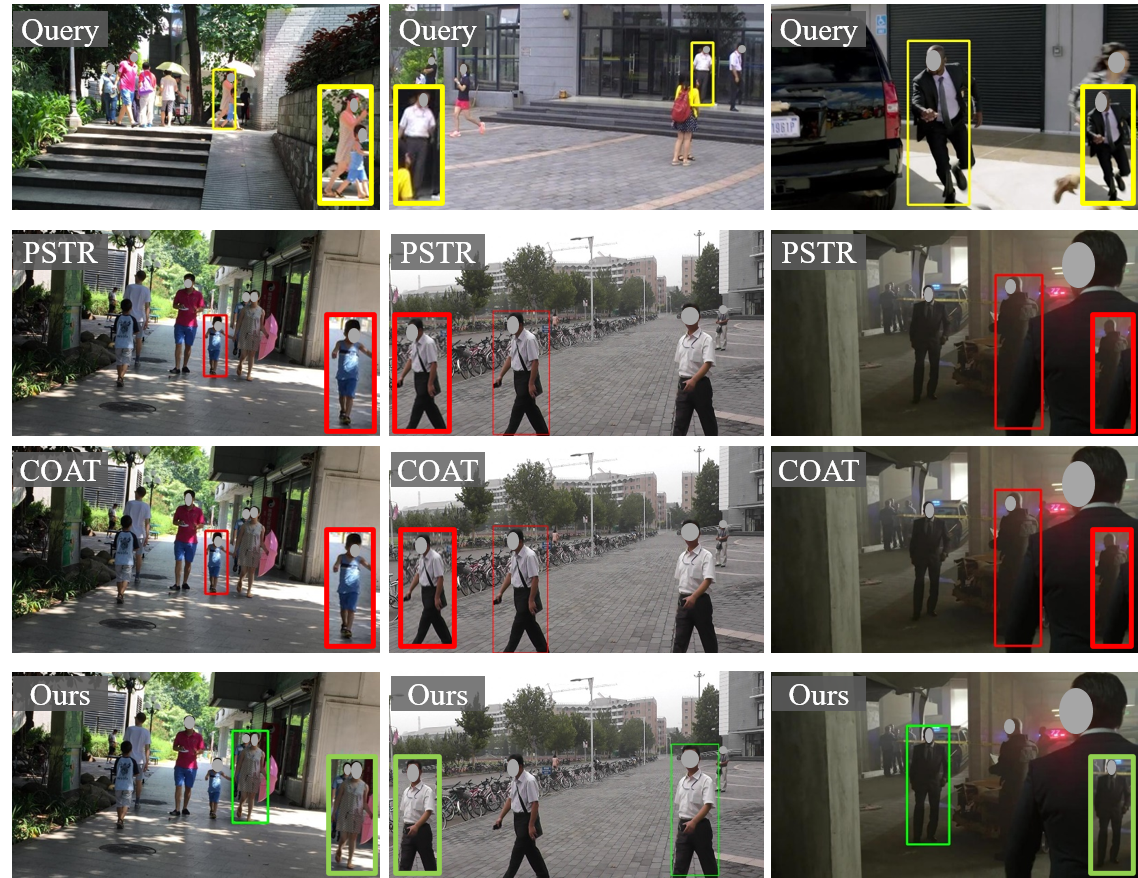} \vspace{-0.6cm}
\caption{Qualitative comparison of our LEAPS with PSTR \cite{Cao_PSTR_CVPR_2022} and COAT \cite{Yu_COAT_CVPR_2022}. The red (green) boxes represent wrong (correct) search results, respectively. Our LEAPS can ignore the inference of surrounding persons (first two columns) and accurately find persons in different scenes (last  column).} \vspace{-0.4cm}
\label{fig:vis}
\end{figure}

\noindent \textbf{Impact of diverse re-id sample matching:}  Table \ref{tab:dsm} shows the impact of diverse re-id sample matching (DRSM) using different re-id heads. Our DRSM presents significant improvements for different re-id heads, especially for our flexible heads (b-d). In addition, the re-id weighting  in DRSM also presents some improvements for different re-id heads. We argue that  diverse sample matching strategy can provide more diverse samples, which are important for  dynamic offset prediction in dynamic RoIAlign and instance-specific re-id feature learning in HIRM.

\noindent \textbf{Impact of cascade structure:}  Table \ref{tab:cascade} shows the impact of cascade stages. We observe that it achieves a good trade-off between accuracy and speed when employing six stages. In addition, we provide the impact of adding the flexible re-id head at several last  stages. It achieves a best performance when adding the re-id heads at last two stages. Therefore, we adopt cascade structure with six stages and only add the re-id heads at last two stages.

\noindent \textbf{Qualitative results:}  Fig. \ref{fig:vis}  compares our LEAPS with the recently proposed PSTR \cite{Cao_PSTR_CVPR_2022} and COAT \cite{Yu_COAT_CVPR_2022}. Our LEAPS can ignore the interference of surrounding persons and provide accurate search results. For example, in first column, PSTR and COAT treat the child around the target woman in query image as the target woman in gallery image.

\section{Conclusion}
In this paper, we present a novel end-to-end person search approach, named LEAPS. Given a set of learnable proposals, our LEAPS directly predicts the person bounding-boxes and corresponding re-id features using a dynamic person search head. In dynamic person search head, we introduce a flexible re-id head to learn discriminative re-id features. In addition, we introduce a diverse sample matching strategy for re-id. With the  ResNet50 backbone, our LEAPS achieves a mAP score of 55.0\% and  top-1 accuracy of 88.9\% at an inference speed of 48 $ms$ on PRW.
Extensive experiments on CUHK-SYSU and PRW datasets demonstrate the superiority of our  LEAPS.  We observe our LEAPS to struggle in heavy occlusions and will explore this direction in future work.

\appendix

\section{More Qualitative Results}

\begin{figure*}[ht]
\includegraphics[width=0.98\linewidth]{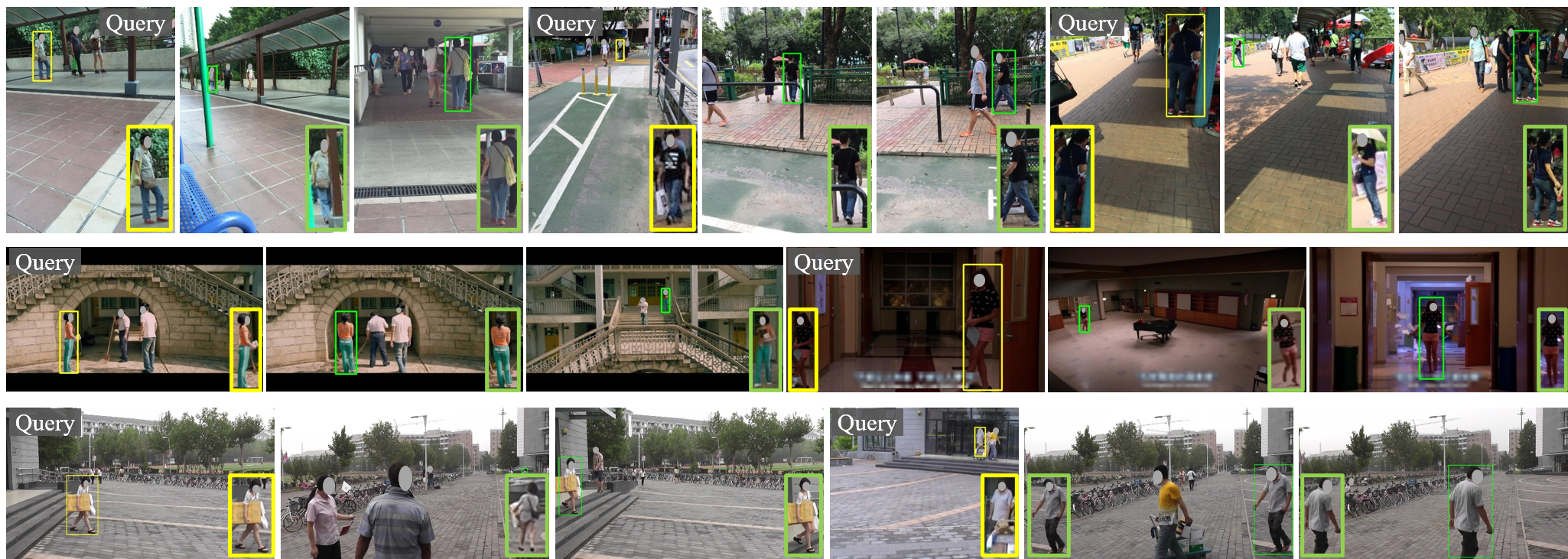}
\caption{Qualitative results on CUHK-SYSU \cite{Xiao_OIM_CVPR_2017} and PRW \cite{Zheng_PRW_CVPR_2017} test sets. The first two rows show  CUHK-SYSU results, while the last row shows PRW results.  For each target person (yellow) in query image, we provide two best search results (green) in gallery images. We add a mask on the person faces for privacy protection.}
\label{fig:viscuhk}
\end{figure*}

Fig. \ref{fig:viscuhk} presents example qualitative results of our LEAPS on both CUHK-SYSU \cite{Xiao_OIM_CVPR_2017} and PRW \cite{Zheng_PRW_CVPR_2017} test sets. Our LEAPS can accurately find the target persons from gallery images under different scenes.

\section{Impact of the number of learnable queries}

Table \ref{tab:num} shows the impact of the number of learnable queries. For a fair comparison with PSTR \cite{Cao_PSTR_CVPR_2022}, we adopt the number of queries as 100, which has the mAP of 55.0\% and the inference speed of 48 $ms$.

\begin{table}[ht]
\footnotesize
\renewcommand{\arraystretch}{1.0}
\begin{center}
\begin{tabular}{|l|ccc|}
\hline
Number & 50    & 100     & 200       \\
\hline \hline
mAP  &   54.5   & 55.0  &  55.2 \\
Time (ms) & 45 & 48  & 51  \\
\hline
\end{tabular}\vspace{-0.5cm}
\end{center}
\caption{Impact of the number of learnable queries.}\vspace{-0.3cm}
\label{tab:num}
\end{table}

{\small
\bibliographystyle{ieee_fullname}
\bibliography{egbib}
}

\end{document}